\documentclass[11pt,a4paper]{article}
\usepackage[hyperref]{AACL-IJCNLP2020}
\usepackage{times}
\usepackage{latexsym}

\usepackage{microtype}

\usepackage{url}

\usepackage{wrapfig}
\usepackage{float}
\usepackage{microtype}
\usepackage{bm}
\usepackage{amsmath,amssymb,amsfonts}
\usepackage{enumerate}
\usepackage[linesnumbered,ruled]{algorithm2e}
\usepackage{subfigure}
\usepackage{graphicx}
\usepackage{multirow}
\usepackage{booktabs}
\usepackage{arydshln}
\usepackage{xcolor}
\usepackage{cases}
\usepackage{color}
\definecolor{bluei}{RGB}{0,150,255}
\definecolor{redi}{RGB}{183,15,18}
\definecolor{greeni}{RGB}{111,139,62}

\aclfinalcopy

\title{High-order Refining for End-to-end Chinese Semantic Role Labeling}

\author{
Hao Fei$^1$, Yafeng Ren$^2$ \and Donghong Ji$^1$\thanks{~~Corresponding author.} \\
1. Department of Key Laboratory of Aerospace Information Security and Trusted Computing,\\
Ministry of Education, School of Cyber Science and Engineering, Wuhan University, China \\ 
2. Guangdong University of Foreign Studies, China \\
\texttt{\{hao.fei,renyafeng,dhji\}@whu.edu.cn}\\
}

\date{}

\begin{document}
\maketitle
\begin{abstract}
Current end-to-end semantic role labeling is mostly accomplished via graph-based neural models.
However, these all are first-order models, where each decision for detecting any predicate-argument pair is made in isolation with local features.
In this paper, we present a high-order refining mechanism to perform interaction between all predicate-argument pairs.
Based on the baseline graph model, our high-order refining module learns higher-order features between all candidate pairs via attention calculation, which are later used to update the original token representations.
After several iterations of refinement, the underlying token representations can be enriched with globally interacted features.
Our high-order model achieves state-of-the-art results on Chinese SRL data, including CoNLL09 and Universal Proposition Bank, meanwhile relieving the long-range dependency issues.
\end{abstract}

\section{Introduction}
Semantic role labeling (SRL), as the shallow semantic parsing aiming to detect the semantic predicates and their argument roles in texts, plays a core role in natural language processing (NLP) community \cite{pradhan-etal-2005-semantic,zhao-etal-2009-semantic,lei-etal-2015-high,XiaL0ZFWS19}.
SRL is traditionally handled by two pipeline steps: predicate identification \cite{SCHEIBLE10} and argument role labeling \cite{pradhan-etal-2005-semantic}.
More recently, growing interests are paid for developing end-to-end SRL,
achieving both two subtasks, i.e., recognizing all possible predicates
together with their arguments jointly, via one single model \cite{he-etal-2018-jointly}.

The end-to-end joint architecture can greatly alleviate the error propagation problem,
thus helping to achieve better task performance.
Currently, the end-to-end SRL methods largely are graph-based neural models,
enumerating all possible predicates and their arguments exhaustively \cite{he-etal-2018-jointly,cai-etal-2018-full,LiHZZZZZ19}.
However, these first-order models that only consider one predicate-argument pair at a time can be limited to short-term features and local decisions, thus being subjective to long-range dependency issues existing at large surface distances between arguments \cite{chen-etal-2019-capturing,lyu-etal-2019-semantic}.
This makes it imperative to capture the global interactions between multiple predicates and arguments.

\begin{figure}[!t]
\centering \includegraphics[width=1.0\columnwidth]{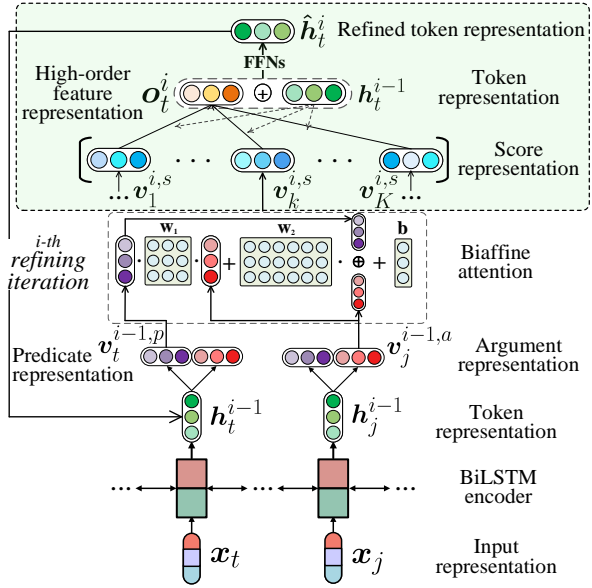}
\caption{
The overview of the graph-based high-order model for end-to-end SRL.
The dotted-line green box is our proposed high-order refining module.
}
\label{encoder}
\end{figure}

In this paper, based on the graph-based model architecture, 
we propose to further learn the higher-order interaction between all predicate-argument pairs by performing iterative refining for the underlying token representations.
Figure \ref{encoder} illustrates the overall framework of our method.
The BiLSTM encoder \cite{hochreiter1997long} first encodes the inputs into the initial token representations for producing predicate and argument representations, respectively.
The biaffine attention then exhaustively calculates the score representations for all the candidate predicate-argument pairs.
Based on all these score representations, our high-order refining module generates high-order feature for each corresponding token via an attention mechanism, which is then used for upgrading the raw token representation.
After total $N$ iterations of the above refining procedure, the information between the predicates and the associated arguments can be fully interacted, and thus results in global consistency for SRL.

On the other hand, most of the existing SRL studies focus on the English language, while there is little work in Chinese, mainly due to the limited amount of annotated data.
In this study, we focus on the Chinese SRL.
We show that our proposed high-order refining mechanism can be especially beneficial for such lower-resource language.
Meanwhile, our proposed refining process is fully parallel and differentiable.

We conduct experimentS on the dependency-based Chinese SRL datasets, including CoNLL09 \cite{hajic-etal-2009-conll}, and Universal Proposition Bank \cite{akbik2015generating,akbik2016polyglot}.
Results show that the graph-based end-to-end model with our proposed high-order refining consistently brings task improvements, compared with baselines, achieving state-of-the-art results for Chinese end-to-end SRL.

\section{Related Work}

\newcite{gildea-jurafsky-2000-automatic} pioneer the task of semantic role labeling, as a shallow semantic parsing.
Earlier efforts are paid for designing hand-crafted discrete features with machine learning classifiers \cite{pradhan-etal-2005-semantic,PunyakanokRY08,zhao-etal-2009-semantic}.
Later, a great deal of work takes advantages of neural networks with distributed features \cite{fitzgerald-etal-2015-semantic,roth-lapata-2016-neural,marcheggiani-titov-2017-encoding,strubell-etal-2018-linguistically}.
On the other hand, many previous work shows that integrating syntactic tree structure can greatly facilitate SRL \cite{marcheggiani-etal-2017-simple,he-etal-2018-syntax,zhang-etal-2019-syntax-enhanced,Crossfei9165903}.

Prior studies traditionally separate SRL into two individual subtasks, i.e., predicate disambiguation and argument role labeling, mostly conducting only the argument role labeling based on the pre-identified predicate \cite{pradhan-etal-2005-semantic,zhao-etal-2009-semantic,fitzgerald-etal-2015-semantic,he-etal-2018-syntax,Fei06295}.
More recently, several researches consider the end-to-end solution that handles both two subtasks by one single model.
All of them employs graph-based neural model, exhaustively enumerating all the possible predicate and argument mentions, as well as their relations \cite{he-etal-2018-jointly,cai-etal-2018-full,LiHZZZZZ19,xia-etal-2019-syntax}.
Most of these end-to-end models, however, are first-order, considering merely one predicate-argument pair at a time.
In this work, we propose a high-order refining mechanism to reinforce the graph-based end-to-end method.

Note that most of the existing SRL work focuses on the English language, with less for Chinese, mainly due to the limited amount of annotated data \cite{xia-etal-2019-syntax}.
In this paper, we aim to improve the Chinese SRL and make compensation of the data scarcity by our proposed high-order model.

\section{Framework}

\paragraph{Task formulation.}

Following prior end-to-end SRL work \cite{he-etal-2018-jointly,LiHZZZZZ19}, we treat the task as predicate-argument-role triplets prediction.
Given an input sentence $S = \{w_1, \cdots, w_n\}$,
the system is expected to output a set of triplets $\mathcal{Y} \in \mathcal{P} \times \mathcal{A} \times \mathcal{R}$,
where $\mathcal{P} = \{p_1, \cdots, p_m\}$ are all possible predicate tokens,
$\mathcal{A} = \{a_1, \cdots, a_l\}$ are all associated argument tokens, and $\mathcal{R}$ are the corresponding role labels for each $a_i$, including a null label $\epsilon$ indicating no relation between a pair of predicate argument.

\subsection{Baseline Graph-based SRL Model}

Our baseline SRL model is mostly from \newcite{he-etal-2018-jointly}.
First, we obtain the vector representation $\bm{x}^{w}_t$ of each word $w_t$ from pre-trained embeddings.
We then make use of the part-of-speech (POS) tag for each word, and use its embedding $\bm{x}^{pos}_t$.
A convolutional neural networks (CNNs) is used to encode Chinese characters inside a word $\bm{x}^{c}_t$.
We concatenate them as input representations: $\bm{x}_t = [\bm{x}^{w}_t;\bm{x}^{pos}_t;\bm{x}^{c}_t]$.

Thereafter, a multi-layer bidirectional LSTM (BiLSTM) is used to encode the input representations into contextualized token representations: $\bm{h}_1,\cdots,\bm{h}_n = \text{BiLSTM}(\bm{x}_1,\cdots,\bm{x}_n)$.
Based on the token representations, we further generate the separate predicate representations and argument representations:
$\bm{v}^p_t=\text{FFN}(\bm{h}_t), \bm{v}^a_t=\text{FFN}(\bm{h}_t)$.
Then, a biaffine attention \cite{dozat2016deep} is used for scoring the semantic relationships exhaustively over all the predicate-argument pairs:
\begin{equation}
\setlength\abovedisplayskip{2pt}
\setlength\belowdisplayskip{2pt}
\label{biaffine attention}
\bm{v}^s(p_{i}, a_{j}) = \bm{v}^p_i \cdot \bm{W}_{1} \cdot \bm{v}^a_j  + \bm{W}_{2} \cdot [\bm{v}^p_i;\bm{v}^a_j] + \bm{b},
\end{equation}
where $\bm{W}_{1}$, $\bm{W}_{2}$ and $\bm{b}$ are parameters.

\paragraph{Decoding and learning.}

Once a predicate-argument pair $(p_{i}, a_{j})$ (i.e., the role label $r\ne\epsilon$) is determined by a softmax classifier, based on the score representation $\bm{v}^s(p_{i}, a_{j})$, the model outputs this tuple $(p,a,r)$.

During training, we optimize the probability  $P_{\theta}(\hat{y}| S)$ of the tuple $y_{(p_{i},a_{j},r)}$ over a sentence $S$:
\begin{equation}
\setlength\abovedisplayskip{3pt}
\setlength\belowdisplayskip{3pt}
\begin{aligned}
\label{Learning}
P_{\theta}(y|S) &= \prod_{p \in \mathcal{P}, a \in \mathcal{A}, r \in \mathcal{R}} P_{\theta}(y_{(p,a,r)}| S) \\
              &= \prod_{p \in \mathcal{P}, a \in \mathcal{A}, r \in \mathcal{R}} \frac{ \phi(p,a,r)}{\sum_{\hat{r} \in R} \phi(p,a,\hat{r})},
\end{aligned}
\end{equation}
where $\theta$ is the parameters of the model and $\phi(p,a,r)$ represents the total unary score from:
\begin{equation}
\setlength\abovedisplayskip{3pt}
\setlength\belowdisplayskip{3pt}
\begin{aligned}
\label{scoring}
\phi(p,a,r) &= \bm{W}_p \text{ReLU}(\bm{v}^p)  + \bm{W}_a \text{ReLU}(\bm{v}^a) \\
 &+ \bm{W}_s \text{ReLU}(\bm{v}^s(p, a)) \,.
\end{aligned}
\end{equation}
The final objective is to minimize the negative log-likelihood of the golden structure:
\begin{equation}
\setlength\abovedisplayskip{3pt}
\setlength\belowdisplayskip{3pt}
\mathcal{L} = - log P_{\theta}(y|S).
\end{equation}

\subsection{Higher-order Refining}

The baseline graph model is a first-order model, since it only considers one predicate-argument pair (as in Eq. \ref{scoring}) at a time.
This makes it limited to short-term and local decisions, and thus subjective to long-distance dependency problem wherever there are larger surface distances between arguments.
We here propose a higher-order refining mechanism for allowing a deep interactions between all predicate-argument pairs.

Our high-order model is shown in Figure \ref{encoder}.
Compared with the baseline model, the main difference lies in the high-order refining module.
Our motivation is to inform each predicate-argument pair with the information of the other rest of pairs from the global viewpoint.
We reach this by refining the underlying token representations $\bm{h}_t$ with refined ones which carry high-order interacted features.

Concretely, we take the baseline as the initiation, performing refinement iteratively.
At the $i$-th refining iteration, we can collect the score representations $\bm{V}^{i,s}=\{\bm{v}^{i,s}_1,\cdots,\bm{v}^{i,s}_K\}$ of all candidate predicate-argument pairs, where $K$ (i.e., $\binom{n}{2}$) are the total combination number of these pairs.
Based on $\bm{V}^{i,s}$, we then generate the high-order feature vector $\bm{o}^{i}_{t}$ by using an attention mechanism guided by the current token representation $\bm{h}^{i-1}_{t}$ for word $w_t$ at last turn, i.e., the ($i$-1)-th iteration:
\begin{equation}
\setlength\abovedisplayskip{3pt}
\setlength\belowdisplayskip{3pt}
\begin{split}
    u^i_k &= \text{tanh}(\bm{W}_3 \bm{h}^{i-1}_{t} + \bm{W}_4 \bm{v}^{i,s}_{k} ), \\
    \alpha^i_k &= \text{softmax}(u^i_k), \\
    \bm{o}^{i}_{t} &=  \begin{matrix} \sum_{k=1}^K \end{matrix} \alpha^i_k \bm{v}^{i,s}_{k} \,,
\end{split}
\end{equation}
where $\bm{W}_3$ and  $\bm{W}_4$ are parameters.
We then concatenate the raw token representation and high-order feature representation together, and obtain the refined token representation after a non-linear projection $\bm{\hat{h}}^{i}_{t}=\text{FFN}([\bm{o}^{i}_{t}; \bm{h}^{i-1}_{t}])$.
Finally, we use $\bm{\hat{h}}^{i}_{t}$ to update the old one $\bm{h}^{i-1}_{t}$.  
After total $N$ iterations of high-order refinement, we expect the model to capture more informative features at global scope and achieve the global consistency.

\section{Experiments}

\subsection{Settings}

Our method is evaluated on the Chinese SRL benchmarks, including CoNLL09\footnote{\url{https://catalog.ldc.upenn.edu/LDC2012T03}} 
and Universal Proposition Bank (UPB)\footnote{\url{https://github.com/System-T/UniversalPropositions}}.
Each dataset comes with its own training, development and test sets.
Precision, recall and F1 score are used as the metrics.

We use the pre-trained Chinese fasttext embeddings\footnote{\url{https://fasttext.cc/}}.
The BiLSTM has hidden size of 350, with three layers.
The kernel sizes of CNN are [3,4,5].
We adopt the Adam optimizer with initial learning rate of 1e-5.
We train the model by mini-batch size in [16,32] with early-stop strategy.
We also use the contextualized Chinese word representations, i.e., ELMo\footnote{\url{https://github.com/HIT-SCIR/ELMoForManyLangs}} and BERT (Chinese-base-version)\footnote{\url{https://github.com/google-research/bert}}.

\begin{table}[!t]
\begin{center}
\resizebox{1.0\columnwidth}{!}{
  \begin{tabular}{lcccc}
\toprule
& \multicolumn{3}{c}{ Arg.}& Prd. \\
\cmidrule(r){2-4}\cmidrule(r){5-5}
& P	&R	&F1	&F1	\\
\midrule
\multicolumn{5}{l}{$\bullet$  \bf  Pipeline method}\\
\newcite{zhao-etal-2009-semantic} &	80.4  &		75.2  &		77.7 &	 -\\
\newcite{bjorkelund-etal-2009-multilingual} &82.4 &		75.1  &		78.6 &	 -\\
\newcite{roth-lapata-2016-neural} &	83.2  &		75.9  &		79.4 &	 -\\
\newcite{marcheggiani-titov-2017-encoding} &84.6 &		 80.4 &		 82.5 &	 -\\
\newcite{he-etal-2018-syntax} &	84.2 &		 81.5 &		 82.8 &	 -\\
\newcite{cai-lapata-2019-semi}$^{\ddagger}$  & \underline{85.4}  &	\underline{84.6}  &	\underline{85.0} &	 -\\
\hline
\multicolumn{5}{l}{$\bullet$  \bf  End-to-end method}\\
\newcite{he-etal-2018-jointly} &  82.6	 &	83.6 &		83.0 &		85.7 \\
\newcite{cai-etal-2018-full} &  84.7 &		 84.0 &		84.3 &		86.0 \\
\newcite{LiHZZZZZ19} & \underline{84.9} &		84.6 &		84.8 &		86.9 \\
\newcite{xia-etal-2019-syntax} & 84.6  & \underline{85.7}  &	\underline{85.1} &	 \underline{87.2} \\
\quad +\texttt{BERT} & 88.0 &	 	89.1  &		88.5 &		89.6 \\
\hdashline
Ours & \bf 85.7 &	\bf 86.2 &	\bf 85.9 &	\bf 88.6 \\
\quad +\texttt{ELMo} &  86.4 &	 87.6 &	 87.1 &	88.9 \\
\quad +\texttt{BERT} & \bf 87.4  &	\bf 	89.3  &	\bf 88.8 &	\bf 90.3 \\
\bottomrule
\end{tabular}
}
\end{center}
  \caption{
  Performances on CoNLL09.
  Results with $^{\ddagger}$ indicates the additional resources are used.
}
  \label{CoNLL09}
\end{table}

\begin{table}[!t]
\begin{center}
\resizebox{0.74\columnwidth}{!}{
  \begin{tabular}{lccc}
\toprule
& P	&R	&F1	\\
\midrule
\newcite{he-etal-2018-jointly} &  64.8 & 	65.3 & 	64.9 \\
\newcite{cai-etal-2018-full} &  65.0 & 	66.4 & 	65.8 \\
\newcite{LiHZZZZZ19} & 65.4 & 	67.2 & 	66.0 \\
\newcite{xia-etal-2019-syntax} & 65.2 & 	67.6 & 	66.1 \\
\hdashline
Ours  &	\underline{67.5} & 	\underline{68.8} & 	\underline{67.9} \\
\quad +\texttt{ELMo} & 68.0 & 	70.6 & 	68.8 \\
\quad +\texttt{BERT} &\bf 70.0	 &\bf 73.0  & \bf	72.4 \\
\bottomrule
\end{tabular}
}
\end{center}
  \caption{
  Performances by end-to-end models for the argument role labeling on UPB.
}
  \label{UPB}
\end{table}

\subsection{Main Results}

We mainly make comparisons with the recent end-to-end SRL models, as well as the pipeline methods on standalone argument role labeling given the gold predicates.
Table \ref{CoNLL09} shows the results on the Chinese CoNLL09.
We first find that the joint detection for predicates and arguments can be more beneficial than the pipeline detection of SRL, notably with 85.1\% F1 score on argument detection by \newcite{xia-etal-2019-syntax}.
Most importantly, our high-order end-to-end model outperforms all these baselines on both two subtasks, with 85.9\% F1 score for argument role labeling and 88.6\%  F1 score for predicate detection.
When the contextualized word embeddings are available, we find that our model can achieve further improvements, i.e., 88.8\% and 90.3\% F1 scores for two subtasks, respectively.

Table \ref{UPB} shows the performances on UPB.
Overall, the similar trends are kept as that on CoNLL09.
Our high-order model still performs the best, yielding 67.9\% F1 score on argument role labeling, verifying its prominent capability for the SRL task.
Also with BERT embeddings, our model further wins a great advance of performances.

\subsection{Analysis}

\begin{figure}[!t]
\centering \includegraphics[width=0.9\columnwidth]{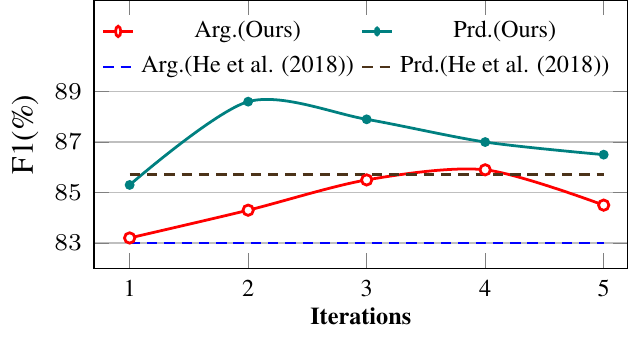}
\caption{
Performances by varying refining iterations.
}
\label{iterations}
\end{figure}

\paragraph{High-order refinement.}
We take a further step, looking into our proposed high-order refining mechanism.
We examine the performances under varying refining iterations in Figure \ref{iterations}.
Compared with the first-order baseline model by \newcite{he-etal-2018-jointly}, our high-order model can achieve better performances for both two subtasks.
We find that our model can reach the peak for predicate detection with total 2 iterations of refinement, while the best iteration number is 4 for argument labeling.

\paragraph{Long-distance dependencies.}

\begin{figure}[!t]
\centering \includegraphics[width=0.9\columnwidth]{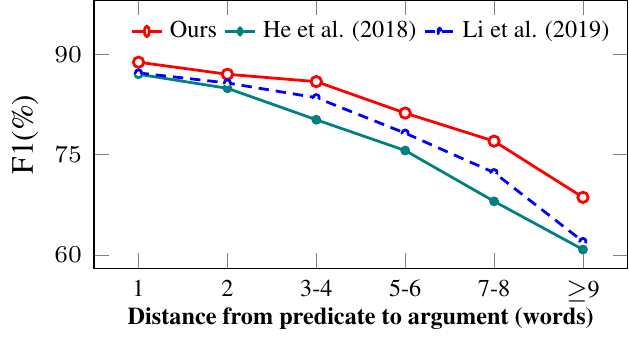}
\caption{
Argument recognition under varying surface distance between predicates and arguments.
}
\label{distance}
\end{figure}

Figure \ref{distance} shows the performances of argument recognition by different surface distance between predicates and arguments.
The overall results decrease when arguments are farther away from the predicates.
Nevertheless, our high-order model can beat against such drop significantly.
Especially when the distance grows larger, e.g., distance $\ge7$, the winning score by our model even becomes more notable.

\section{Conclusion}

We proposed a high-order end-to-end model for Chinese SRL.
Based on the baseline graph-based model, our high-order refining module performed interactive learning between all predicate-argument pairs via attention calculation.
The generated higher-order featured token representations then were used to update the original ones.
After total $N$ iterations of refinement, we enriched the underlying token representations with global interactions, and made the learnt features more informative.
Our high-order model brought state-of-the-art results on Chinese SRL data, i.e., CoNLL09 and Universal Proposition Bank, meanwhile relieving the long-range dependency issues.

\section{Acknowledgments}

We thank the anonymous reviewers for their valuable and detailed comments.
This work is supported by the National Natural Science Foundation of China (No. 61772378, No. 61702121), 
the National Key Research and Development Program of China (No. 2017YFC1200500), 
the Research Foundation of Ministry of Education of China (No. 18JZD015), 
the Major Projects of the National Social Science Foundation of China (No. 11\&ZD189),
the Key Project of State Language Commission of China (No. ZDI135-112) 
and Guangdong Basic and Applied Basic Research Foundation of China (No. 2020A151501705).

\bibliography{aacl-ijcnlp2020}

\bibliographystyle{acl_natbib}

\end{document}